# Quadruped Robot Simulation Using Deep Reinforcement Learning

Nabeel Ahmad Khan Jadoon
Department of Industrial System
Engineering
Asian Institute of Technology
Bangkok, Thailand
st123321@ait.asia

*Abstract*—I present a novel reinforcement learning method to train the quadruped robot in a simulated environment. The idea of controlling quadruped robots in a dynamic environment is quite challenging and my method presents the optimum policy and training scheme with limited resources and shows considerable performance. The report uses the raisimGymTorch open-source library and proprietary software RaiSim for the simulation of ANYmal robot. My approach is centered on formulating Markov decision processes using the evaluation of the robot walking scheme while training. Resulting MDPs are solved using a proximal policy optimization algorithm used in actor-critic mode and collected thousands of state transitions with a single desktop machine. This work also presents a controller scheme trained over thousands of time steps shown in a simulated environment. This work also sets the base for early-stage researchers to deploy their favorite algorithms and configurations.

## I. INTRODUCTION

The development of methods to address the challenge of perceptive locomotion on unorganized terrain for legged robots has made great advances in recent years. The advancements range from Mini-Cheetah [1], Spot-min [2], and other legged robotic platforms that provide different controlling schemes and software. For instance, Pybullet, ROS, Gazebo, and others have gained significant attention over the last 5 years.

This work presents the unique implementation of control policy on ANYmal [1] robot in the RaiSim physic engine and uses deployment of C++ and packages to build environments and constraints. RaiSim is a physics engine developed by RaiSim Tech Inc which provides state-of-the-art flexibility, accuracy, and speed for simulating robotic systems. It also offers minimum dependency on STL and Eigen libraries thus making it an interesting physic engine as compared to others benchmarks [3]. The general interface of the software is presented in Fig. 1 which runs on the Unity framework and RaiSim server in South Korea.

The choice of legged robots is indeed diverse but the locomotory performance of the human-like robots is still far behind their natural counterparts. Better locomotion performance in terms of speed, energetic efficiency, and obstacle negotiation skills, is achieved with multi-legged systems. A paramount example is Boston Dynamics' Spot robot, a direct successor of Big Dog [4], of which unfortunately no scientific publications are available.

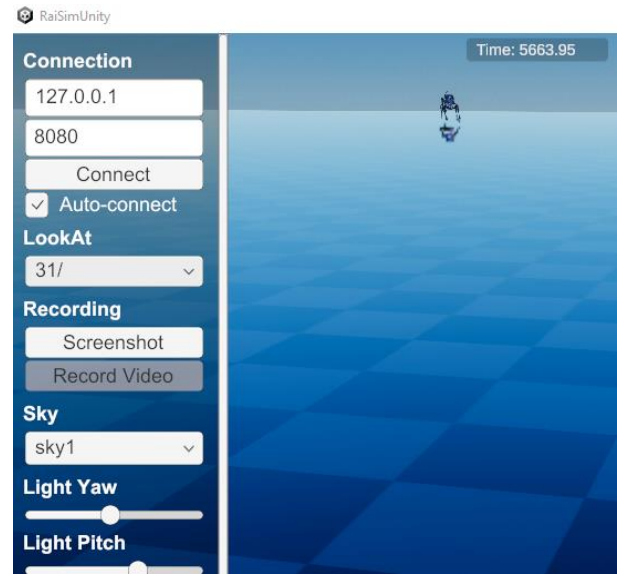

Fig.1 RaiSim Simulator

The main aim of this project to use the ANYmal robot is that it is a rugged quadrupedal platform developed for autonomous operation in challenging environments. ANYmal combines outstanding mobility with dynamic motion capability that enables it to climb large obstacles and dynamic runs. That is why the implementation of the reinforcement scheme presented in this report shows outstanding locomotion thanks to the innovative design of ANYmal as shown in Fig 2.

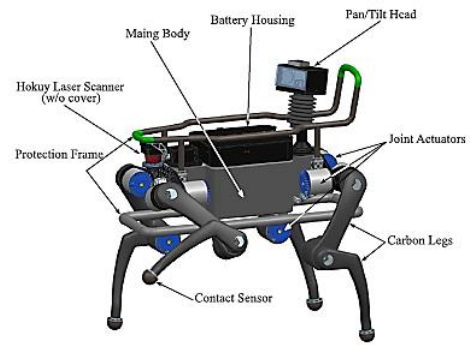

Fig 1. Main components of ANYmal robot

This robot also provides the URDF file extension to be exported for researchers and hence was used in this project inside the RaiSim simulator where collision bodies, contact

Warch Full Series of Training:  https://youtu.be/KRpQKGhey0o
Code Available in my Github repository: https://github.com/nabeeljadoon/ANYmalRobotSimulation_RaiSim.git

forces, and angles are customizable. Fig 3 shows the imported URDF 3D CAD file.

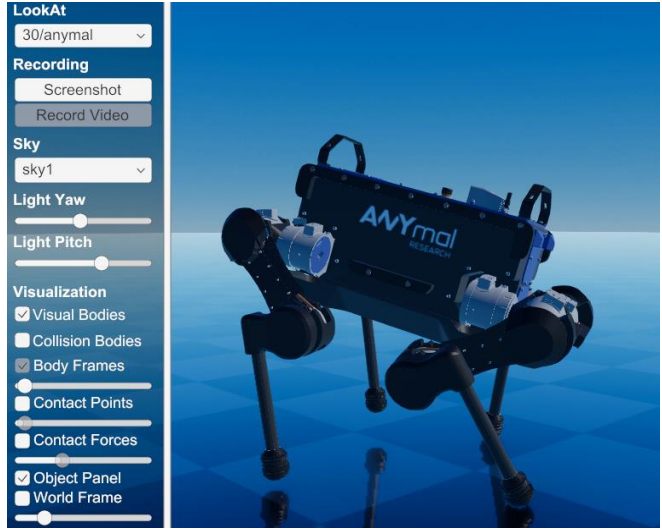

Fig 3. ANYmal inside RaiSim

## II. LITERATURE AND RELATED WORK

### A. Selecting a Trajectory optimization for legged robots

Within the present literature, two broad families of approaches have become most prominent: 1) Model-based Trajectory optimization (TO) and 2) Deep reinforcement learning (DRL). The past works addressing terrain-aware locomotion problems have predominantly used model-based approaches, such as those employing deterministic optimization techniques [5], in conjunction with other heuristics [6] for impedance based-motion of base and feet. However, more recent developments that use DRL [7], [8] tend to relax many of the modeling assumptions, but they still need precise physical simulation to function well in real systems.

### B. Deep Reinforcement on Legged Robots

Deep reinforcement learning (RL) has recently been used to train bipeds and quadrupeds to learn end-to-end finish controllers. A general RL method is given by Hwangbo et al. [30] for training joint angle controllers from the robot's base and joint states. The authors suggest building a model of ANYmal's actuation dynamic [9] from real-world data that can then be used in simulation to enable the transfer of learned rules to the actual world. A central pattern generator (CPG) layer that generates a standard walking gait pattern for the feet is introduced to the action space in the work by Miki et al. [10] using a similar learning technique. Utilizing a LIDAR-based reconstruction of the surroundings and proprioception, the policy then learns to alter the joint angles and CPG phase to change the gait.

Like this, Lee et al. [11] discover a strategy that alters the phase and shift of CPG functions, which dictate the foot trajectories given to a model-based controller to generate joint angle control. Ji et al [12] proposal for learning a control policy through RL and a state estimation network with supervised learning aims to predict state variables, such as foot contact states and linear velocity of the base, that cannot be measured on the actual robot but are available in simulation and are crucial for learning robust policies.

This project presents deep reinforcement learning which combines a neural network with RL algorithms to learn the value function approximations [13], policies [14], and imitating tasks such as learning to walk schemes. The proximal policy optimization (PPO) [15] was used and presented results on the simulated ANYmal robot inside the RaiSim simulator.

The following sections will shed light on reinforcement learning parameters, MDP state, actions, and reward functions. Moreover, the result section will also show the simulation while training to quickly determine the policy evaluation.

## III. METHODOLOGY

The goal of this project is to showcase the imitation and learning scheme for quadruped robots using deep reinforcement learning on ANYmal. The description of MDP components and Pytorch framework is defined in the following sections. The general flow for DRL is given in Fig 4.

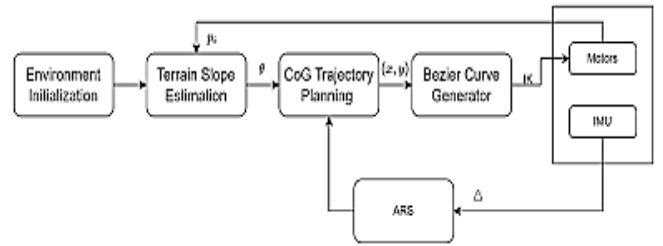

Fig 4. Adaptive Motion Skill Learning (Walk)

For this robot, the controlling scheme shown in Fig 5 the command is given by the controller or pc, and resulting state measurements and estimations are done from the actual dynamics of the robot.

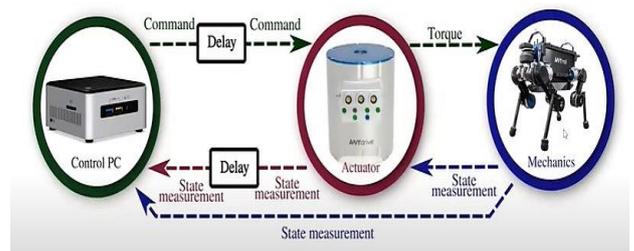

Fig 5. RL Control Architecture for ANYmal Robot

By using the model-free methods, we can generate control policies without any abstraction in modeling complex dynamics. However, model-free algorithms require several trials and errors to obtain a better policy and hence become impractical to train. Imitating robot behavior like self-righting and locomotion are presented in a simulated environment for this project using proximal policy optimization (PPO). Moreover, a learning-based state estimator that is learned in parallel to the training of the robot is presented which even works in complex, unobservable contact conditions.

## A. Overview of the WorkFlow

For this project, the tasks were decomposed to primarily three behaviors, locomotion and standing up, and righting behaviors. This division is to tackle the difficulty of training a single policy that can manifest all of the necessary behaviors. It is because previous research [16] showed that a policy trained to perform multiple tasks shows unpredicted behaviors such as frequent slippages and conservative postures. On the other hand, learning three behaviors separately simplifies the cost function design and enables us to troubleshoot each control policy separately on the simulated robot and can also be leveraged for the real system.

While training is only performed in simulation, every behavior is individually tested on the real robot and trained on it. The behavior selection and a height estimator are then used employing the collection of previously trained behaviors. The quick contact solver introduced in [17] and the data-driven actuator model [18] make up the simulated environments for learning. These tools effectively produce high-fidelity samples. For learning, the PPO method combines the estimator, and stochastic policies are employed during training, however, to achieve more consistent behavior upon deployment, the variances are lowered to 0.

## B. State Spaces (Feature Selection)

The given table 1 represents a reliable set of states and the existing state estimation framework of ANYmal relies on base pose and twist, and angular encoders for the joint states. For instance, Kinematic contact limitations are included by using the placements of feet on the ground. The predicted base states from TSIF become unreliable when a foot slips or when all four feet lose contact with the ground, which is likely to occur when ANYmal falls, as the position and linear velocity drift over time.

| Function | Data |
|---|---|
| Self-Righting Policy | Gravity vector ($e_g$) <br> Base angular velocity in body frame ($\omega_{IB}^B$) <br> Joint positions ($\phi_j$) <br> Joint velocities ($\dot{\phi}_j$) <br> History of joint position error & velocity <br> Previous joint position targets ($a_{t-1}$) |
| Standing Up Policy | Base linear velocity in body frame ($v_{IB}^B$) <br> State space of the Self-Righting policy |
| Locomotion Policy | Velocity commands <br> Estimated base height ($h_e$) <br> State space of the Standing up policy |
| Behavior Selector | Previous action (one-hot vector) <br> State space of the Locomotion policy |
| Height Estimator | Gravity vector ($e_g$) <br> Joint positions <br> Joint velocities <br> History of joint position errors & velocities |
| Actuator Model | Desired joint position <br> History of joint position errors & velocities |

Table 1: Definition of State Space

## C. Symbols and Cost Terms

The linear combination of cost terms and symbols for the ANYmal robot is shown in Table 2. For joint position cost, the minimum angle difference between two angular positions is denoted by R xR → [0, pi].

All the equations and representations are mentioned in the given table.

| Symbols | |
|---|---|
| $\phi_{jslim}$ | maximum joint speed |
| $I_c$ | index set of the contact points |
| $I_{c,f}$ | index set of the foot contact points |
| $I_{c,in}$ | index set of the self-collision points |
| $i_{c,n}$ | impulse of the $n$th contact |
| $g_i$ | gap function of the $i$th contact |
| $p_{f,i}$ | position of the $i$th foot |
| $\tau$ | vector of joint torques |
| $|\cdot|$ | cardinality of a set or $l_1$ norm |
| $\|\cdot\|$ | $l_2$ norm |
| $\hat{\phantom{x}}$ | target value |
| **Cost Terms** | |
| Angular velocity | $c_\omega = K(|\omega_{IB}^B - \hat{\omega}_{IB}^B|, \alpha_a)$ |
| Linear velocity | $c_v = K(|v_{IB}^B - \hat{v}_{IB}^B|, \alpha_l)$ |
| Height | $c_h = 1.0$ if $h < 0.35$, otherwise 0 |
| Joint position | $c_{jp} = d_\phi(\phi_j, \hat{\phi}_j)$ |
| Orientation | $c_o = \|[0, 0, -1]^T - e_g\|$ |
| Torque | $c_\tau = \|\tau\|^2$ |
| Power | $c_{pw} = \sum_{i=0}^{12} max(\dot{\phi}_{j,i}\tau_i, 0)$ |
| Joint acceleration | $c_a = \sum_{i=0}^{12} \|\ddot{\phi}_i\|^2$ |
| Joint speed | $c_{js} = \sum_{i=0}^{12} max(\dot{\phi}_{jslim} - |\dot{\phi}_i|, 0)^2$ |
| Body impulse | $c_{bi} = \sum_{n \in I_c \setminus I_{c,f}} \|i_{c,n}\|/(|I_c| - |I_{c,f}|)$ |
| Body slippage | $c_{bs} = \sum_{n \in I_c} \|v_{c,n}\|^2/|I_c|$ |
| Foot slippage | $c_{fs} = \sum \|v_{f,i}\|$ <br> $\forall i$ s.t. $g_i = 0, i \in I_{f,c}$ |
| Foot clearance | $c_{fc} = \sum (p_{f,i} - 0.07)^2\|v_{f,i}\|$ <br> $\forall i$ s.t. $g_i > 0, i \in I_{f,c}$ |
| Self collision | $c_{cin} = |I_{c,in}|$ |
| Action difference | $c_{ad} = \|a_{t-1} - a_t\|^2$ |

Table 2: Symbols and Cost Terms for ANYmal robot

## D. Robots Behaviors and Selector Scheme

**Locomotion behavior** is to follow given command schemes composed of desired forward velocity, lateral velocity, and turning rate or yaw rate. A major part of this report was focused on achieving the best locomotion behavior of the robot. The defined min and max velocities are 0.4 and 1.2 m/s respectively. The cost function penalizes the velocity tracking errors (cw and cv), foot motions (cfc and cfs), and constraint violation. The equation is mentioned in the appendix section of this report.

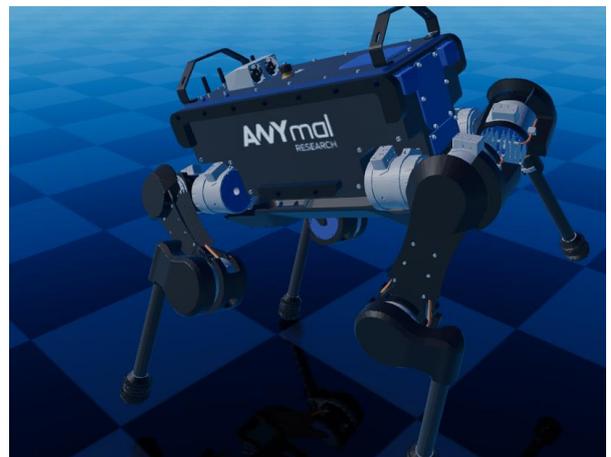

Fig 6. Locomotion position of the robot inside RaiSim

Similarly, the **Standing up behavior** and **self-righting behavior** were constructed and configured. For standing up, an arbitrary sitting configuration was considered, and the high cost was additionally introduced. While self-righting behavior is to regain an upright base pose from an arbitrary configuration and re-position joints to the sitting configurations which is designed so that ANYmal has all feet on the ground for safe stand-up maneuver.

The following algorithm depicts the behavior selection control scheme with PPO in actor-critic configurations.

**Algorithm 1 Training Behavior Selector**

Initialize $\theta, \psi$ randomly
for $i = 0, 1, ..., N$ do
  for $t = 0, 1, ..., T$ do
    if $i > N_w$ then     ▷ $N_w$ = Warm-up period
      Use the estimated height $h_\psi(s_t)$
    Sample action $a_t \sim \pi_\theta(a|s_t)$
    Excute the corresponding behavior
    Collect state $s_t$, action $a_t$, and reward $r_t$
    Collect true height $h_t$.
    Append a $s_t$-$h_t$ pair into the replay memory
  Sample $K$ pairs from the replay memory
  Update $\psi$ by minimizing $\sum_{j=0}^{K}||h_j - h_\psi(s_j)||^2$
  Update $\theta$ using **PPO Actor-Critic Mode**

*E. Algorithm and Binaries*

RaisimGymTorch wraps a C++ environment file as a Python library using pybind11. The calling of the setup configuration environment compiled the designed environments inside the library and stored them in respective directories.
The detailed configuration of the code structure is given in the following Fig 7.

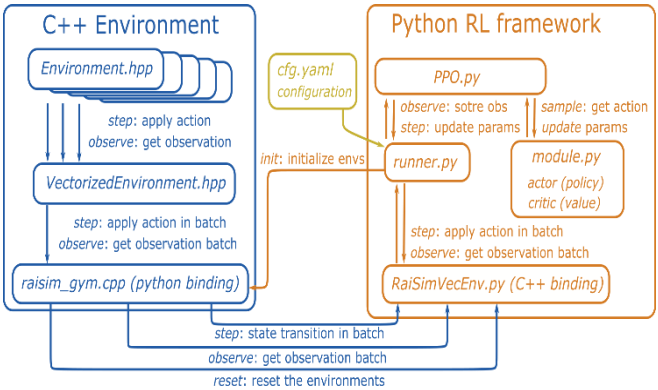

Fig 7. Code Structure and Wrapping

   *a. PPO Algorithm*

In this project, the PPO algorithm with actor-critic mode was developed to achieve the best possible locomotion for a quadruped robot. It is considered a state-of-the-art algorithm [19]. The control policy algorithm for this robot is shown in Fig 8. PPO builds on the TRPO foundation which is Trust Region Policy Optimization somewhat authors claimed that it is similar to the policy gradient method. PPO has two parts comprising 1) PPO penalty and, 2) PPO clip.
PPO -variant with adaptive KL penalty [19] is a popular technique that provides penalty over different constraints set for robot locomotion. The discussion of this technique is not in \the scope of this project. The alternative technique that outperforms the penalty-based variant and is simpler to implement is called PPO-Variant with the clipped objective [18].

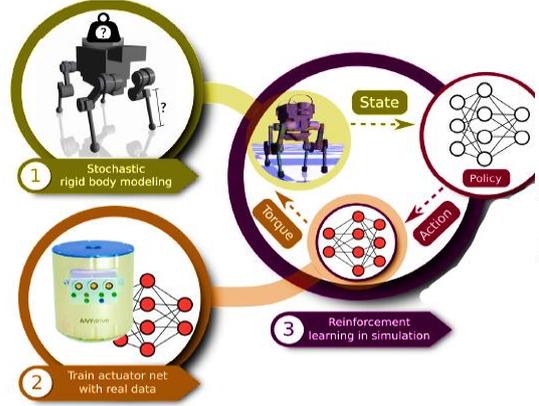

Fig 8. Control Policy

It is indeed a cumbersome process to change the penalties over time, we can simply restrict the range within which the policy can change. Hence, the advantages achieved by updates outside the clipping range are not used for updating purposes and provide merit to stay relatively close to the existing policy. The equation for the clip function is mentioned below and it's called **surrogate advantage.**

$$\mathcal{L}_{\pi_\theta}^{CLIP}(\pi_{\theta_k}) = \mathbb{E}_{\tau \sim \pi_\theta}\left[\sum_{t=0}^{T}\left[\min\left(\rho_t(\pi_\theta, \pi_{\theta_k})A_t^{\pi_{\theta_k}}, \text{clip}(\rho_t(\pi_\theta, \pi_{\theta_k}), 1-\epsilon, 1+\epsilon)A_t^{\pi_{\theta_k}}\right)\right]\right]$$

Similarly, the important sampling ratio used in this project is given by the following formula:

$$\rho_t(\theta) = \frac{\pi_\theta(a_t \mid s_t)}{\pi_{\theta_k}(a_t \mid s_t)}$$

As the variables (1-ϵ)·A and (1+ϵ)·A are independent, the gradient is equal to 0. As samples beyond the trusted region are effectively discarded as a result, excessively large updates are discouraged. As a result, we don't expressly limit the policy update itself and instead choose to disregard benefits brought on by excessive policy deviation. Like before, we can just make the modifications using an optimizer like ADAM.

In this variant of PPO, the surrogate advantage is clipped. If the updated policy deviates from the original one by more than ϵ, the sample yields a gradient of 0. The mechanism avoids overly large updates of the policy, retaining it within a trusted region. Refer to below Fig 9. for a clear understanding.

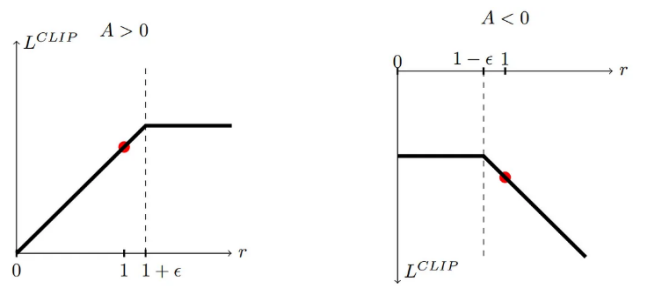

Fig 9. Clipped Objective Variant of PPO

**Algorithm 5** PPO with Clipped Objective

Input: initial policy parameters $\theta_0$, clipping threshold $\epsilon$
for $k = 0, 1, 2, \ldots$ do
  Collect set of partial trajectories $\mathcal{D}_k$ on policy $\pi_k = \pi(\theta_k)$
  Estimate advantages $\hat{A}_t^{\pi_k}$ using any advantage estimation algorithm
  Compute policy update
  $$\theta_{k+1} = \arg\max_\theta \mathcal{L}_{\theta_k}^{CLIP}(\theta)$$
  by taking $K$ steps of minibatch SGD (via Adam), where
  $$\mathcal{L}_{\theta_k}^{CLIP}(\theta) = \mathop{\mathbb{E}}_{\tau \sim \pi_k}\left[\sum_{t=0}^{T}\left[\min\left(r_t(\theta)\hat{A}_t^{\pi_k}, \text{clip}(r_t(\theta), 1-\epsilon, 1+\epsilon)\hat{A}_t^{\pi_k}\right)\right]\right]$$
end for

## IV. IMPLEMENTATION

The controller was tested inside the simulated environment to train a high-speed locomotion task with an increased velocity of 1.6 m/s and maintain constant velocity for 10 sec. The maximum torque computed is 40 Nm and the maximum joint velocities are 12 rad/s.

### A. Addition of Noise

The noise added up to 0.2 m/s of noise to the linear velocity, 0.25 rad/s to the angular velocity, and 0.5 rad/s to the angular velocity to duplicate the noisy observation from the real robot. During training in a simulation (RaiSim), the joint locations and 0.05 rad to the joint velocities are adjusted.

### B. Training Control Policies in Simulation

The policy network links the joint position targets with the current observation and the joint state history. In rigid body modeling, the actuator network converts the joint state history to the joint torque. The generalized coordinate q and the generalized velocity u make up the robot's state. The joint position error, which is the present position (phi) subtracted. from the joint position goal (phi*), and the joint velocity w make up a joint's state as shown in Fig 10.

## V. EXPERIMENTAL RESULTS

To obtain the required results for this project, the knowledge from DRL lab work was leveraged to implement the simulation while training to check whether the robot is learning something or not.

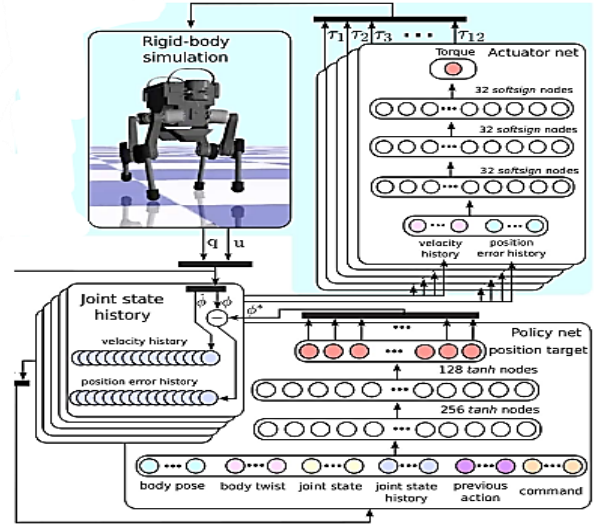

Fig 10. Policy network Mapping with MLP

### A. Simulation While Training

For the simulation environment, a single desktop machine was used and on a physics engine called RaiSim training and testing schemes were performed. The data during the training was directed to the Tensor-flow board to visualize the different graphs and will be shown in the following sections. For example, the following two iteration schemes showed parameters and average learning rewards.

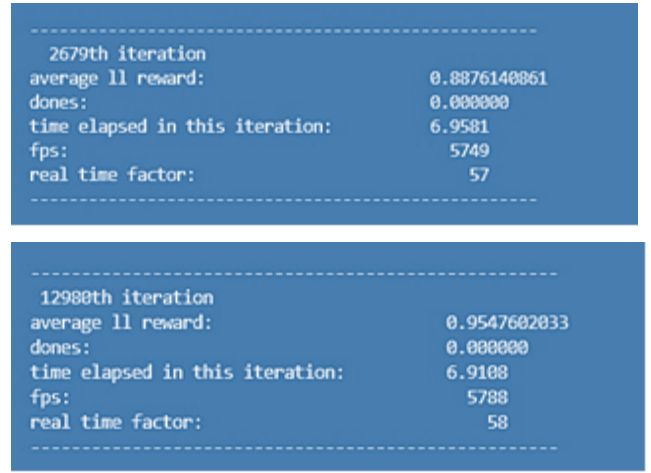

Fig 11. Iterations during training.
*Training: https://youtube.com/shorts/Ufa-ZafTNMU*

The robot was trained for 13K iterations and the estimated training time was 7 hours, using GPU on a single machine. During training, the collected iteration schemes were captured for visualizing different graphs. The final training video is given in the appendix section.

These training weights are imperative to check the behavior of a robot inside the simulated environment. Due to the limitation and scope of the project, the maximum transitions were gathered around 13K iterations. However, the higher number of transitions shows a better policy in terms of deploying or transferring simulation to the real world. During training, the reward iterations were collected and shown in Fig 12.

![Terminal output showing reward values]

Fig 12. Reward vs iterations

Results show the stabilization and locomotion of the robot. Below graph 1 depicts the surrogate advantage function used in the PPO algorithm for whole iterations and relates to the learning rate shown in Graph 2. The important factor revealed by Graph 3 is the value function given at 200 iterations states. The policy was trained to accumulate for 200 episodes and then logged the data while simulating. This helped to identify the hyperparameter tuning for optimization.

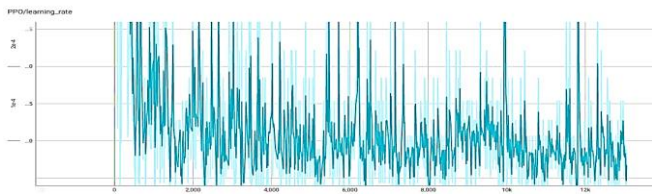

Graph 1: Surrogate Advantage Function

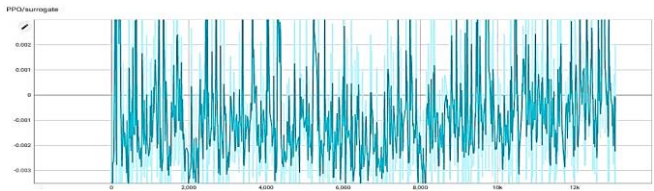

Graph 2: PPO Learning rate over 13K iterations

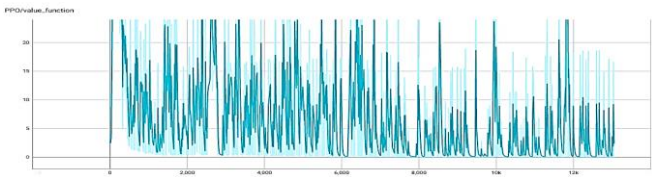

Graph 3: PPO Value function

## VI. CONCLUSION

The work scheme presented in this paper was trained in a few hours in a simulation inside the RaiSim environment for 30K iterations and policy was visualized during training right after 200 iteration steps. The learned locomotion policies ran faster and with higher precision while utilizing low resources, less energy, and less torque.

The proposed method is not limited to well-known and straightforward situations. The findings in this study regard as a first step toward thorough locomotion controllers for tough and adaptable-legged robots.

## REFERENCES

[1] M. Hutter, C. Gehring, D. Jud, A. Lauber, C. D. Bellicoso, V. Tsounis, J. Hwangbo, K. Bodie, P. Fankhauser, M. Bloesch, et al., "ANYmal-a highly mobile and dynamic quadrupedal robot," in IEEE/RSJ Int. Conf. on Intelligent Robots and Systems. IEEE, 2016, pp. 38–44.

[2] B. Dynamics, "Spot Mini Robot, www.bostondynamics.com/spot.

[3] https://leggedrobotics.github.io/SimBenchmark/

[4] M. Raibert, K. Blankespoor, G. Nelson, and R. Playter, "BigDog, the rough-terrain quadruped robot," in Proceedings of the 17th World Congress, pp. 10823–10825, 2008.

[5] S. Kuindersma, R. Deits, M. Fallon, A. Valenzuela, H. Dai, F. Permenter, T. Koolen, P. Marion, and R. Tedrake, "Optimization-based locomotion planning, estimation, and control design for the atlas humanoid robot," Autonomous Robots, pp. 429–455, 2016.

[6] X. B. Peng, G. Berseth, K. Yin, and M. Van De Panne, "DeepLoco: Dynamic Locomotion Skills Using Hierarchical Deep Reinforcement Learning," ACM Trans. Graph., pp. 41:1–41:13, 2017.

[7] C. D. Bellicoso, F. Jenelten, C. Gehring, and M. Hutter, "Dynamic Locomotion Through Online Nonlinear Motion Optimization for Quadrupedal Robots," IEEE Robot. Automat. Lett., vol. 3, no. 3, pp. 2261–2268, 2018.

[8] M. Hutter et al., "Animal - a highly mobile and dynamic quadrupedal robot," in 2016 IEEE/RSJ Int. Conf. Intell. Robots Syst., 2016, pp. 38–44.

[9] J. Hwangbo, J. Lee, A. Dosovitskiy, D. Bellicoso, V. Tsounis, V. Koltun, and M. Hutter, "Learning agile and dynamic motor skills for legged robots," Science Robotics, vol. 4, no. 26, 2019

[10] Z. Li, X. Cheng, X. B. Peng, P. Abbeel, S. Levine, G. Berseth, and K. Sreenath, "Reinforcement Learning for Robust Parameterized Locomotion Control of Bipedal Robots," Proceedings - IEEE Int. Conf. Robot. Automat., vol. 2021-May, pp. 2811–2817, 2021.

[11] https://towardsdatascience.com/proximal-policy-optimization-ppo-explained-abed1952457b.

[12] F. Farshidian, M. Neunert, A. W. Winkler, G. Rey, J. Buchli, An efficient optimal planning and control framework for quadrupedal locomotion,2017 IEEE International Conference on Robotics and Automation, 93–100 (IEEE, 2017).

[13] R. Tedrake, T. W. Zhang, H. S. Seung, Stochastic policy gradient reinforcement learning on a simple 3d biped, 2004 IEEE/RSJ International Conference on Intelligent Robots and Systems, 2849–2854 (IEEE, 2004).

[14] J. Schulman, F. Wolski, P. Dhariwal, A. Radford, O. Klimov, Proximal policy optimization algorithms, arXiv preprint arXiv:1707.06347 (2017). 24. N. Heess, S. Sriram, J. Lemmon, J. Merel, G. Wayne, Y. Tassa, T. Erez, Z. Wang, A. Eslami, M. Riedmiller, others, Emergence of locomotion behaviors in rich environments, arXiv preprint arXiv:1707.02286 (2017).

[15] Michael Bloesch et al. State estimation for legged robots consistent fusion of leg kinematics and IMU. Robotics,17:17–24, 2013.

[16] Michael Bloesch et al. The Two-State Implicit Filter Recursive Estimation for Mobile Robots. IEEE Robotics and Automation Letters, 3(1):573–580, 2018.

[17] Dian-sheng Chen, Jun-mao Yin, Yu Huang, Kai Zhao, and Tian-miao Wang. Hopping-righting mechanism analysis and design of the mobile

robot. Journal of the Brazilian Society of Mechanical Sciences and Engineering, 35(4):469–478, 2013.

[18] https://towardsdatascience.com/proximal-policy-optimization-ppo-explained-abed1952457b.

Appendix: Report

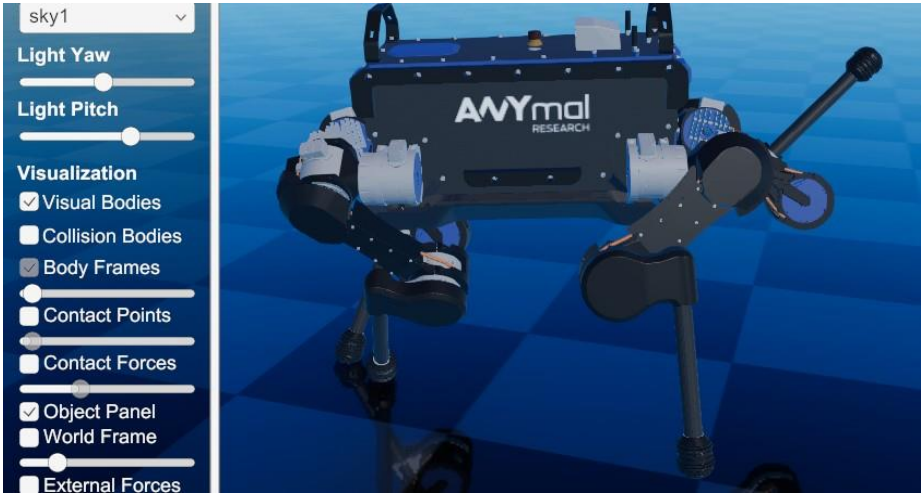

AnyMAL Error

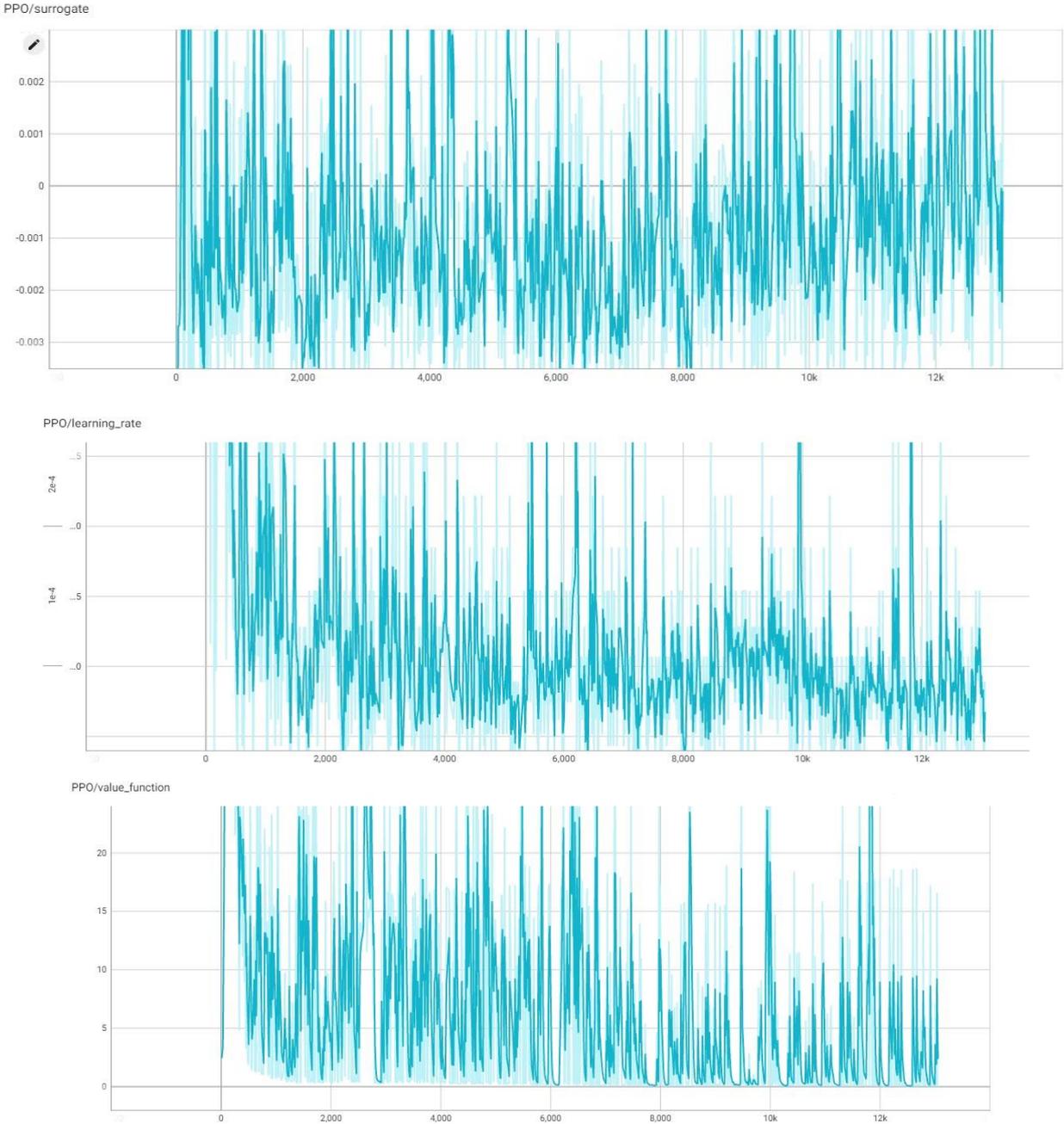

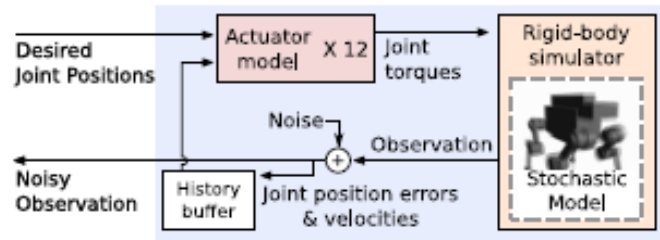

Simulation Model Scheme

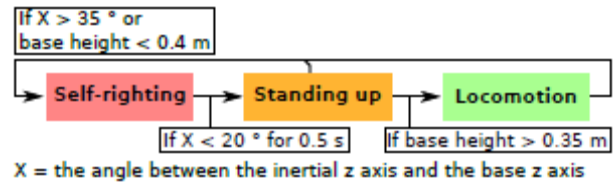

Angle Behavior Selection